\documentclass[11pt,a4paper]{article}
\ifdefined\pdfobjcompresslevel
\fi
\usepackage[margin=1in]{geometry}
\usepackage[T1]{fontenc}
\usepackage[utf8]{inputenc}
\usepackage{lmodern}
\usepackage{amsmath,amssymb}
\usepackage{graphicx,booktabs,pifont}
\usepackage[numbers,square,sort&compress]{natbib}
\usepackage[hidelinks]{hyperref}
\graphicspath{{./}}
\AddToHook{env/table/begin}{\setlength{\belowcaptionskip}{8pt}}
\AddToHook{env/table*/begin}{\setlength{\belowcaptionskip}{8pt}}

\newsavebox{\tablebox}
\newcommand{\fitTable}[1]{%
  \sbox{\tablebox}{#1}%
  \ifdim\wd\tablebox>\linewidth
    \resizebox{\linewidth}{!}{\usebox{\tablebox}}%
  \else
    \usebox{\tablebox}%
  \fi}

\title{Disentangle to create: Architecture-level disentangled diffusion model for few-shot Chinese font generation}
\author{%
Jie Li$^{1,2}$ \and Suorong Yang$^{1,3,*}$ \and
Jian Zhao$^{4}$ \and Furao Shen$^{1,2,*}$}
\date{}
\begin{document}
\maketitle
\begin{center}
\small
$^1$State Key Laboratory for Novel Software Technology, Nanjing University, Nanjing, China\\
$^2$School of Artificial Intelligence, Nanjing University, Nanjing, China\\
$^3$Department of Computer Science and Technology, Nanjing University, Nanjing, China\\
$^4$School of Electronic Science and Engineering, Nanjing University, Nanjing, China\\[0.5em]
\texttt{lijie2002@smail.nju.edu.cn}, \texttt{sryang@smail.nju.edu.cn}\\
\texttt{jianzhao@nju.edu.cn}, \texttt{frshen@nju.edu.cn}\\
$^*$Corresponding authors: Suorong Yang and Furao Shen.
\end{center}
\begin{abstract}
Few-shot Chinese font generation aims to synthesize new characters in a target style using only a handful of reference images. Achieving accurate content rendering and faithful style transfer requires effective disentanglement between content and style. However, most existing approaches achieve only feature-level disentanglement and allow the generator to re-entangle these features, leading to content distortion and degraded style fidelity. To address this, we propose the Architecture-Level Disentangled Diffusion Model (ALD-Font) with an asymmetric design. The U-Net backbone takes SimSun-style glyphs as content templates, concatenated with noisy latents to construct a spatial canvas. Style features extracted from target-style images are injected as modulation signals to guide generation. Based on theoretical validation of disentanglement effectiveness, we adopt a parameter-efficient fine-tuning (PEFT) strategy that updates only the style-related modules. This allows the model to better adapt to new styles while avoiding overfitting to the reference images' content under the few-shot setting. Additionally, we train a Background Noise Removal (BNR) module in the pixel space to remove background noise in complex stroke regions. We further introduce the Grey and OCR metrics to evaluate the content quality of generated characters. Experimental results show that ALD-Font achieves significantly higher style fidelity while maintaining comparable content accuracy to existing state-of-the-art methods. The implementation is available at \url{https://github.com/JayLee1002/ALD-Font}.
\end{abstract}
\section{\label{Xsec1-1}
Introduction }
\label{sec:intro}

Chinese font generation integrates linguistic expression with {aesthetic} design to support personalized branding, historical media authenticity, and digital restoration, promoting cultural creativity and artistic exploration in modern digital contexts \citep{fontgen-survey,fontgen-survey2}. However, unlike English letters, Chinese characters have complex structures and exist in large quantities. For instance, the GB2312 standard includes 6763 commonly used characters, and the GB18030 standard covers over 27,000 \citep{gb1,gb2}. Manually designing Chinese fonts is extremely time-consuming and labor-intensive.
To this end, deep generative models have been increasingly adopted to synthesize Chinese characters in novel styles with minimal reference samples~\citep{cg-gan,mx,diff-font,msdfont,fontify}.
Despite achieving promising results, they struggle to preserve content faithfully while ensuring high-quality style consistency.
\begin{figure}[!htbp]

    \centering
    \includegraphics[width=.90\linewidth,height=0.68\textheight,keepaspectratio]{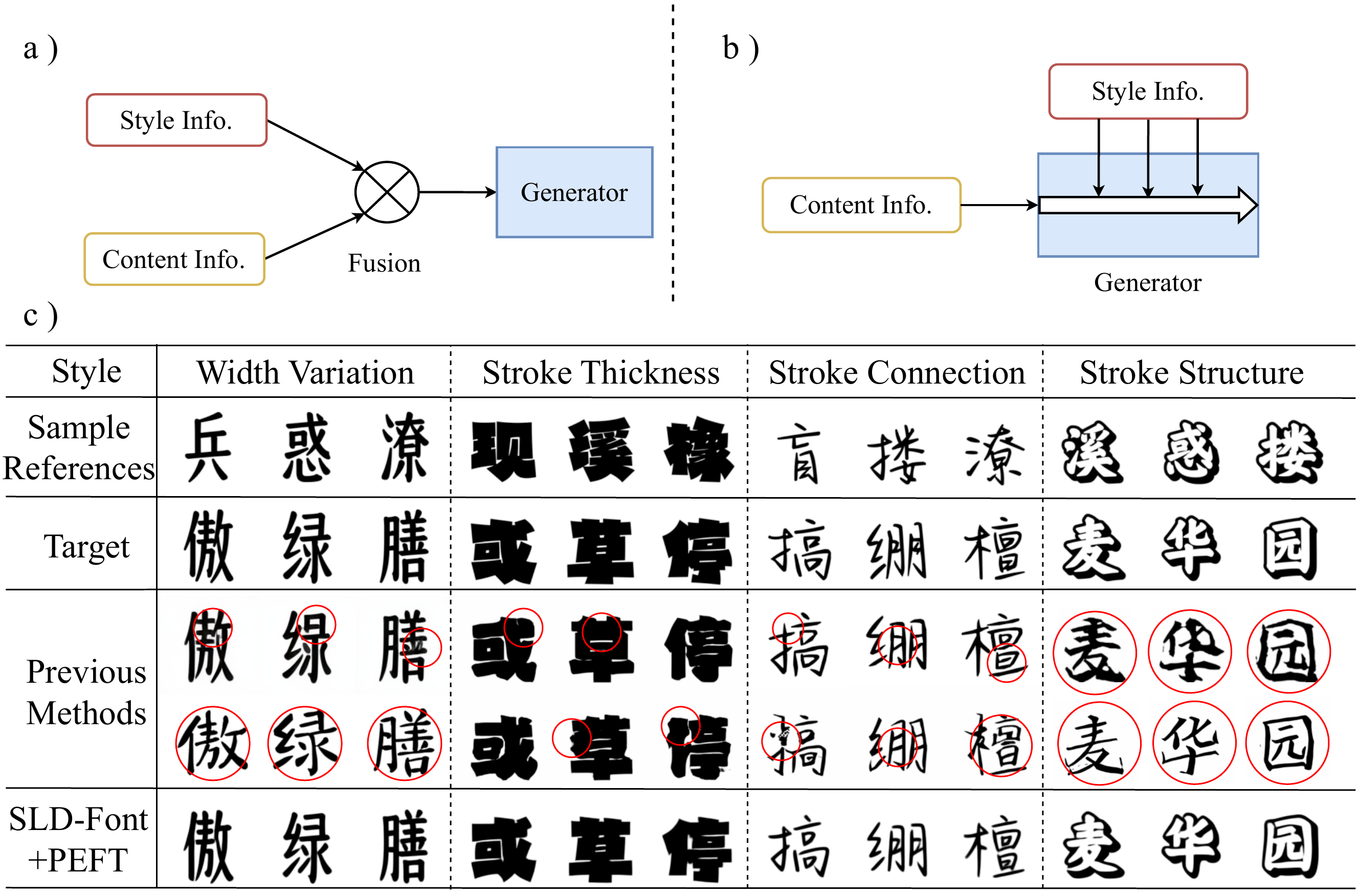}
    
\caption{ Existing methods, as illustrated in (a), fuse style and content as generator input, resulting in feature-level disentanglement; (b) uses content as a spatial canvas and style as guidance, achieving architecture-level disentanglement.
As demonstrated in (c), our method produces Chinese characters with noticeably higher visual quality.
    \label{fig:intro}}
\end{figure}
\begin{figure}[!htbp]

    \centering
    \includegraphics[width=1.0\linewidth,height=0.68\textheight,keepaspectratio]{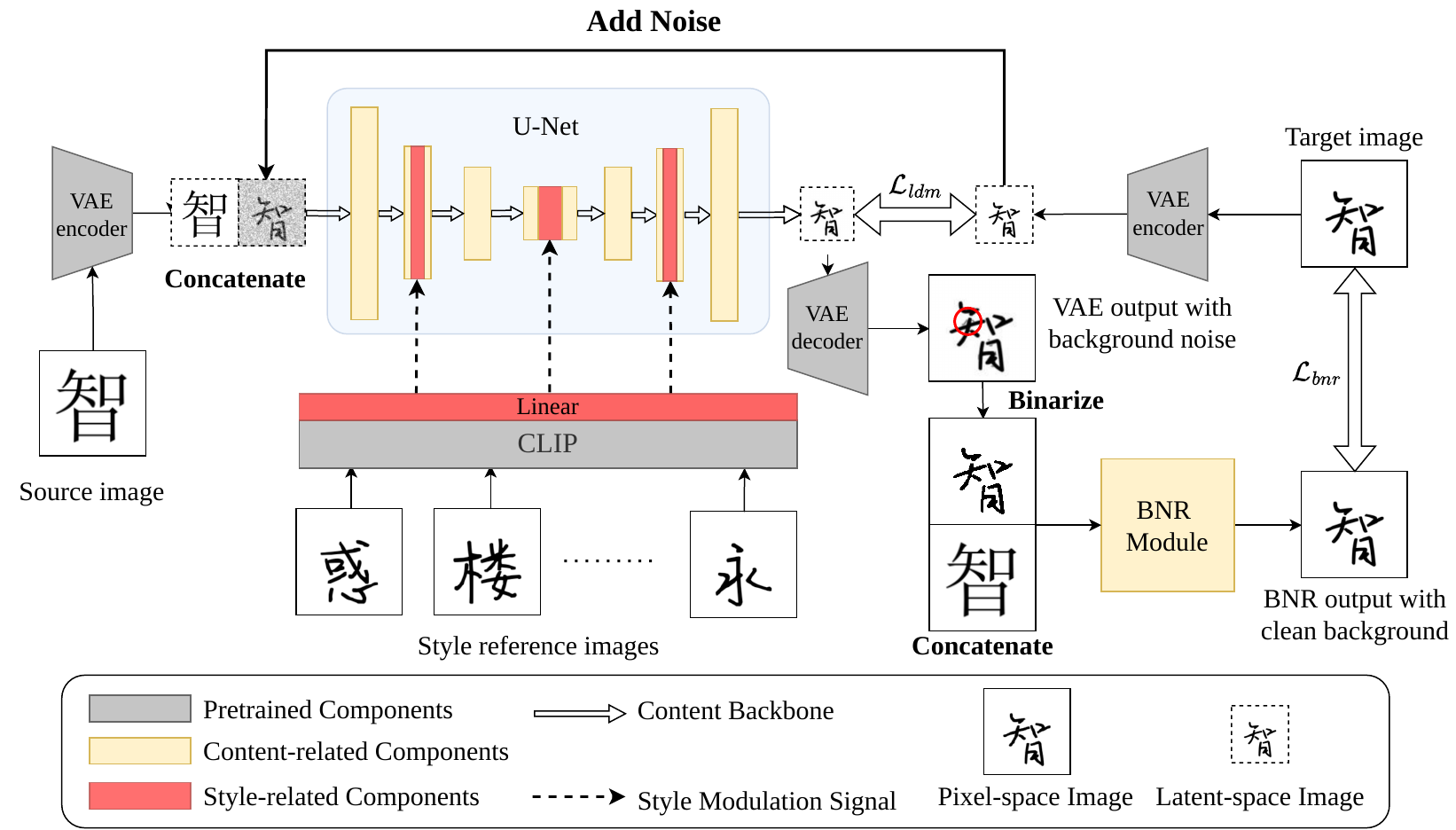}
    
\caption{Both source and target images are encoded into the latent space by a VAE encoder. Gaussian noise is added to the target latent during diffusion, and the resulting noisy latent is concatenated with the source latent and fed into the U-Net. The yellow and red parts denote content-related and style-related components, respectively. Content-related components process the source-guided spatial stream to preserve glyph structure, while style-related components receive CLIP-extracted reference features and inject style modulation through cross-attention in Transformer blocks. The U-Net output is decoded by the VAE, binarized, and further refined in pixel space by the BNR module to produce clean and accurate glyphs.\label{fig:overview}}
\end{figure}
Unlike style transfer \citep{style-transfer,stylediff}, which focuses on transferring color and texture, font generation demands much higher content fidelity, as even slight deviations can lead to content errors.
More importantly, the style of Chinese characters is typically conveyed through attributes, such as width, stroke thickness, connectivity, and structure, as shown in Fig.~{\ref{fig:intro}}(c), making content-style disentanglement a key determinant of visual quality and usability.
While existing frameworks based on VAEs \citep{savae}, GANs \citep{cg-gan,lf,mx,ntf}, Transformer \citep{xmp-font,fontify}, or diffusion models \citep{msdfont,fontdiffuser,diff-font} attempt to separate content and style using dual-encoder designs, they predominantly rely on a symmetric fusion paradigm as in Fig.~{\ref{fig:intro}}(a). In such designs, content and style features are fused into a joint representation before being processed by the generator. Such feature-level disentanglement inevitably requires re-entanglement during the generation, where stylistic updates inadvertently distort the character's structural integrity. As shown in Fig.~{\ref{fig:intro}}(c), the entangled designs produce characters with distorted structures or incomplete adaptation to the target style, e.g., incorrect strokes, inconsistent width variation, erratic stroke thickness, and disjointed stroke  connections.

To address this challenge, we propose an \textbf{A}rchitecture-\textbf{L}evel \textbf{D}isentangled Diffusion Model for few-shot Chinese font generation (ALD-Font).
To overcome the content sensitivity inherent in Chinese characters, ALD-Font employs an asymmetric architecture in which content and style information are processed through different pathways, thereby achieving architecture-level disentanglement.
As illustrated in Figs.~{\ref{fig:intro}}(b) and {\ref{fig:overview}}, our generator is formulated as a content-anchor-driven architecture. We redefine the U-Net backbone as a ``Spatial Canvas'': by using the source image as a template and concatenating it with the latent input, we establish a persistent content anchor. This ensures that the global geometric skeleton remains invariant throughout the diffusion process. Concurrently, style information is extracted from multiple reference images via a CLIP-based encoder and injected through cross-attention as latent modulation signals. This asymmetric modulation, where content provides the spatial constraint and style provides the stylistic modulation, effectively isolates the two components into distinct parameter subspaces.
This architectural disentanglement provides a critical foundation for parameter-efficient fine-tuning (PEFT). During adaptation, by strategically freezing content-related parameters and updating only style-related ones, ALD-Font facilitates rapid stylistic migration, effectively mitigating ``structural collapse''.
Finally, recognizing that latent diffusion often introduces micro-artifacts at high-frequency stroke edges during the conversion between pixel space and VAE latent space, we incorporate a pixel-space Background Noise Removal (BNR) module. By leveraging binarized guidance and content information, the BNR module refines the original output image to meet the rigorous {aesthetic} standards of Chinese typography.

Traditional metrics such as $\ell_1$ and Structural Similarity Index (SSIM) \citep{ssim} mainly measure pixel-level similarity and may not fully capture the structural and semantic requirements of font generation. To address this limitation, we introduce two tailored metrics for content accuracy evaluation: Gray, which measures background noise via grayscale histogram comparison, and an OCR-based accuracy metric to evaluate character recognizability.
Under the challenging Unseen Chars with Unseen Fonts (UCUF) setting, ALD-Font improves SSIM from 0.356 to 0.394 compared with previous methods. Based on architecture-level disentanglement, ALD-Font+PEFT further increases SSIM to 0.436, while sacrificing only 0.005 in OCR accuracy, demonstrating effective style adaptation with minimal impact on content integrity.
We summarize our contributions as follows:
\begin{itemize}
    \item We propose ALD-Font, which reorganizes latent diffusion and conditioning mechanisms into an asymmetric architecture for few-shot Chinese font generation. The U-Net processes content information as a spatial canvas, while style information provides modulation signals via cross-attention, improving content accuracy and style fidelity through architecture-level disentanglement.

\item To the best of our knowledge, we are the first to introduce PEFT into few-shot font generation based on architecture-level disentanglement. By fine-tuning only style-related parameters identified by gradient sensitivity analysis, ALD-Font enables more accurate style capture while mitigating content overfitting.

\item To mitigate artifacts introduced by the VAE, particularly in regions with complex strokes, we design a Background Noise Removal (BNR) module that operates in pixel space to suppress such noise.

\item In addition to traditional evaluation metrics, we introduce two content-level indicators, Gray and OCR. Compared with SOTA methods using various metrics, our approach achieves much higher style similarity while maintaining comparable content accuracy.
\end{itemize}

\section{\label{Xsec2-2}
Related work}

\subsection{\label{Xsec3-2.1}
Few-shot Chinese font generation}

Font generation has evolved from manual design to automatic image synthesis using deep generative models. Early methods focused on single style transfer with fully supervised learning, but these approaches were not scalable to large character sets \citep{fontgen-survey,i2i,zi2zi}. Recent advances have shifted toward few-shot font generation (FFG), which aims to synthesize new fonts from a limited number of reference images by disentangling style and content representations. Early methods used VAEs \citep{vae2013,vae2014}, whereas current methods are mainly based on GAN \citep{gan}, Transformer \citep{transformer}, and diffusion models \citep{ddpm,ddim,sde}.

VAE-based methods attempt to model content-style separation in latent space. SA-VAE \citep{savae} introduces a pairwise optimization strategy to enable style inference from only one or a few reference samples. EMD \citep{emd} proposes an architecture  composed of style and content encoders, allowing transfer to unseen styles and contents.
GAN-based models typically fuse content and style in a single generative stream. AGISNet \citep{agisnet} utilizes two collaboratively working decoders to generate the glyph shape simultaneously. MX-Font \citep{mx} and MX-Font++ \citep{mxfont++} innovate by employing multiple localized experts to automatically extract diverse local styles, while LF-Font \citep{lf} introduces a low-rank factorization mechanism to decompose style into  component and style factors. DG-Font \citep{dg-font} applies feature deformation skip connections with displacement-guided deformable convolution. NTF \citep{ntf} treats font generation as a continuous transformation task using a neural transformation field with an iterative style refinement strategy to align content features across fonts and optimize the reference style representation. Component-aware approaches further introduce structural supervision, such as CF-Font \citep{cffont} with a Content Fusion Module, CG-GAN \citep{cg-gan} with component-aware learning, DA-Font \citep{dafont} with dual-attention component--relation modeling, and Fs-Font \citep{fsfont} with cross-attention alignment between content regions and local styles. Self-supervised signals are also explored, including CLIP-Font \citep{clipfont} with text-semantic supervision and EdgeFont \citep{edgefont} with multi-scale edge guidance.
XMP-Font \citep{xmp-font} employs a Transformer jointly conditioned on glyphs and stroke labels, with a self-supervised cross-modality encoder for multi-scale content--style modeling. Fontify \citep{fontify} reformulates glyph synthesis as context-aware inpainting with concatenated content--style pairs and random block masking.
Diffusion-based methods have recently shown strong potential for FFG due to stable training and high visual fidelity. Diff-Font \citep{diff-font} and HFH-Font \citep{hfh} introduce component-aware conditioning to guide glyph generation. FontDiffuser \citep{fontdiffuser} improves local detail with multi-scale content aggregation and style contrastive refinement. MSD-Font \citep{msdfont} and QT-Font \citep{qt-font} further explore structured glyph representations through staged generation and sparse quadtree encoding.

However, despite evolving backbone architectures, these methods mainly improve feature extraction, feature fusion, or generator design and thus remain limited to feature-level disentanglement. Moreover, many approaches rely on additional prior information to improve generation quality, such as manually annotated component structures in LF-Font, MX-Font, and component-aware methods, stroke-related representations in stroke-guided methods, pretrained visual-semantic representations in CLIP-based methods, or latent diffusion priors in diffusion-based approaches. These additional priors can improve specific aspects of font generation, but they do not explicitly separate the architectural pathways of content preservation and style adaptation. Therefore, existing methods still lack architecture-level disentanglement, which is crucial for preserving character integrity and enabling controllable style transfer.

\subsection{\label{Xsec4-2.2}
Parameter efficient fine-tuning}

Parameter-efficient fine-tuning (PEFT) aims to adapt large pretrained models by updating only a small set of parameters, thereby reducing memory and storage requirements and mitigating overfitting with limited downstream data. Early adapter-based methods insert compact trainable modules between frozen Transformer layers, showing that task adaptation can be achieved without modifying most pretrained weights \citep{adapter}. Prompt- and prefix-based approaches instead optimize continuous task representations: Prefix-Tuning \citep{prefix-tuning} prepends trainable key-value vectors to the attention layers, Prompt Tuning \citep{prompt-tuning} learns soft input prompts, and P-Tuning v2 \citep{ptuningv2} extends deep prompt tuning across model scales and tasks. Other methods directly select small subsets of existing parameters, such as BitFit \citep{bitfit}, which only tunes bias terms, while LoRA \citep{lora} learns low-rank residual updates for selected linear layers. Recent PEFT studies further move beyond a fixed adaptation form. PARA \citep{para} adjusts hidden representations with prompt-aware lightweight vector generators, and Structure-Learnable Adapter Fine-Tuning \citep{slaft} automatically searches for adapter insertion and activation structures. These methods reveal complementary PEFT design choices, including inserted modules, learned prompts, selective parameter tuning, representation adjustment, structure search, and adaptive low-rank updates.

PEFT has also become important for diffusion-based generation, especially in personalization and controllable synthesis. Textual Inversion \citep{textual-inversion} freezes a text-to-image diffusion model and learns only a new token embedding from a few reference images, demonstrating that compact trainable representations can encode new visual concepts. Custom Diffusion \citep{mcc} makes multi-concept customization more efficient by updating selected conditioning parameters rather than the full model. DiffFit \citep{difffit} adapts large diffusion models by tuning only bias terms and lightweight scaling factors, and T2I-Adapter \citep{t2iadapter} keeps the original diffusion model frozen while learning small adapters for external control signals. These works show that generative models can acquire new concepts, domains, or controls through limited trainable components. Our method first achieves architecture-level disentanglement and uses PEFT to specialize the style-related parameters from a small reference set. This combination significantly improves style-related metrics while maintaining content accuracy.

\section{\label{Xsec5-3}
Methodology}

As shown in Fig.~{\ref{fig:overview}}, we propose ALD-Font, an Architecture-Level Disentangled Diffusion Model for effective few-shot Chinese font generation. ALD-Font is an asymmetric latent diffusion model that builds a spatial canvas from SimSun templates and noisy latents, with style modulation guiding generation. Leveraging this disentanglement, we further adopt parameter-efficient fine-tuning (PEFT) to improve adaptability to unseen styles by fine-tuning only style-related parameters. A Background Noise Removal (BNR) module is introduced at the pixel level to correct errors caused by latent--pixel conversion.

\subsection{\label{Xsec6-3.1}
Preliminaries on latent diffusion models}

A diffusion model learns to generate data by reversing the gradual noising process, step by step transforming random noise back into structured data \citep{ddpm,sde}.
ALD-Font is built upon the LDMs \citep{ldm}, which significantly reduce computational cost by performing the diffusion process in a compressed VAE \citep{vae,savae} latent space while preserving high-quality generation. Given an image $x$ in pixel space, it is first encoded into latent space by the VAE encoder as $z_0=\mathcal{E}(x)$. The forward process progressively adds noise to $z_0$ until its distribution approaches a standard Gaussian distribution $z_T \sim \mathcal{N}(\mathbf{0}, \mathbf{I})$:
\begin{align}
z_t &= \sqrt{\alpha_t} z_{t-1} + \sqrt{1 - \alpha_t} \varepsilon, \quad \varepsilon \sim \mathcal{N}(\mathbf{0}, \mathbf{I}), \\
z_t &= \sqrt{\bar{\alpha}_t} z_0 + \sqrt{1 - \bar{\alpha}_t} \varepsilon, \quad \varepsilon \sim \mathcal{N}(\mathbf{0}, \mathbf{I}),\label{add_noise}
\end{align}
where $\alpha_t$ is a time-dependent hyperparameter, $\bar{\alpha}_t=\Pi_{i=1}^t \alpha_i$ and $t \in \{1, 2, \dots, T\}$, with $T$ denoting the total number of diffusion steps.
The reverse process starts by sampling pure noise $z_T \sim \mathcal{N}(\mathbf{0}, \mathbf{I})$ and gradually denoises it to obtain $z_0$. Each step from $z_t$ to $z_{t-1}$ is conditioned on $z_0$ and calculated as:
\begin{align}
&q(z_{t-1} \mid z_t, z_0) = \mathcal{N}(z_{t-1}; \tilde{\mu}_t(z_t, z_0), \tilde{\beta}_t \mathbf{I}), \label{reverse} \nonumber \\
& \tilde{\mu}_t(z_t, z_0) := \frac{\sqrt{\bar{\alpha}_{t-1}}(1 - \alpha_t)}{1 - \bar{\alpha}_t}  z_0 + \frac{\sqrt{\alpha_t}(1 - \bar{\alpha}_{t-1})}{1 - \bar{\alpha}_t} z_t, \nonumber \\
& \tilde{\beta}_t := \frac{1 - \bar{\alpha}_{t-1}}{1 - \bar{\alpha}_t}  (1 - \alpha_t),
\end{align}
The latent variable $z_0$ is then passed through the VAE decoder $\mathcal{D}$ to obtain the reconstructed image in pixel space, denoted as $x = \mathcal{D}(z_0)$.

To implement the reverse process described in Eq.~{\eqref{reverse}}, LDMs use the U-Net model $\hat{z}_\theta$ to predict $z_0$ gradually. Following \cite{msdfont,dreambooth}, we adopt the strategy of directly predicting $z_0$, rather than predicting noise as in the original DDPMs \citep{ddpm}. Noise prediction improves diversity, whereas directly predicting $z_0$ ensures a more deterministic process that better preserves content consistency in font generation. Given conditioning input $c$, the training objective of LDMs at time step $t$ is defined as:
\begin{align}
    \mathcal{L}_{\mathrm{ldm}}=\mathbb{E}_{t,x,\varepsilon} [\parallel z_0 - \hat{z}_\theta(z_t,c) \parallel_2^2],
\end{align}
The U-Net receives and processes conditioning information via cross-attention in Transformer blocks \citep{transformer,ldm}.

\subsection{\label{Xsec7-3.2}
Architecture-level disentangled diffusion model}

ALD-Font achieves architecture-level disentanglement through an asymmetric architecture built upon the U-Net backbone. The overall framework is illustrated in Fig.~{\ref{fig:overview}}. We formulate few-shot font generation as an image-to-image translation task conditioned on a few reference images in the target style, aiming to transform source font glyphs into the target style while preserving content accuracy. SimSun is selected as the source font due to its wide availability across devices and its clear and well-defined character structure. Given a pair $(x, y)$, where $x$ is the SimSun character and $y$ is its corresponding character in the target style, we also provide a set of $N$ target-style reference images ${r_1, r_2, \dots, r_N}$ to provide style information.

For content information, a pre-trained VAE encoder is used to encode $x$ and $y$ into the latent representations $z_{\mathrm{x}}$ and the initial latent representation $z_{\mathrm{y}}^0$. During training, we sample $t \sim \text{Uniform} ({1, \dots, T})$ and add noise to $z_{\mathrm{y}}^0$ following Eq.~{\eqref{add_noise}}:
\begin{align}
    z_{\mathrm{y}}^t
    =\sqrt{\bar{\alpha}_t} z_{\mathrm{y}}^0
    + \sqrt{1 - \bar{\alpha}_t} \varepsilon,
    \quad \varepsilon \sim \mathcal{N}(\mathbf{0}, \mathbf{I}),
\end{align}
Following \citep{sr3}, we concatenate the SimSun-style content image $z_{\mathrm{x}}$ and the noisy target image $z_{\mathrm{y}}^t$ along the channel dimension as the input to the U-Net. The content image of a Chinese character clearly preserves its structural information. Using it as a template provides sufficient guidance for generating structurally accurate results. Thus, the U-Net is designated as the backbone for processing content information.

For style information, we use a pre-trained CLIP model \citep{clip} to extract the style representation from each reference image, denoted as $s_i = \text{CLIP}(r_i)$. This results in a style matrix $s \in \mathbb{R}^{N \times d}$, where $d$ is the dimensionality of the style embedding. The style features $s$ are injected into the U-Net through cross-attention in Transformer blocks, allowing style conditioning during generation. This process resembles text prompts in text-to-image generation, except that our CLIP model treats all reference images equally, making the order of reference images irrelevant. Given the misalignment between CLIP's pretraining objective and font generation \citep{clip2021,clip}, we retrain the final projection layer to retain general visual representations while adapting to our task.

We adopt the strategy of predicting $z_0$, and the loss function is defined as:
\begin{align}
    \hat{z}_{\mathrm{y}}^0
    &= \hat{z}_{\theta}(\text{concat}(z_{\mathrm{y}}^t, z_{\mathrm{x}}), s),
    \label{predict_x0}\\
    \mathcal{L}_{\mathrm{ldm}}
    &= \mathbb{E}_{t,(x,y),\{r_1,r_2,\dots,r_N\},\varepsilon}
    \left[ \left\| z_{\mathrm{y}}^0-\hat{z}_{\mathrm{y}}^0 \right\|_2^2 \right].
    \label{loss}
\end{align}
Here, $z_{\mathrm{y}}^t$ is used as the denoising state in the standard diffusion training process rather than as an additional conditioning signal. The conditioning information is provided by the SimSun glyph latent $z_{\mathrm{x}}$ and the reference style embedding $s$.

Let $\theta = \theta_{\mathrm{c}} \cup \theta_{\mathrm{s}}$ denote the model parameters, corresponding to the content and style pathways highlighted in yellow and red in Fig.~{\ref{fig:overview}}, respectively.
By assigning content information to provide a spatial canvas and style information to deliver latent modulation signals, ALD-Font achieves architecture-level content-style disentanglement. Specifically, this disentanglement refers to the separation of information sources and input pathways, rather than complete statistical independence between content and style. Different from feature-level disentanglement methods that concatenate or fuse encoded content and style features before generation, ALD-Font avoids mixing the two types of information before they enter the generator. Style propagation in later layers is therefore part of the intended guidance mechanism for style transfer. In Section~{\protect\ref{Xsec8-3.3}}, we further identify different parts of the U-Net related to content and style through gradient sensitivity analysis.

\subsection{\label{Xsec8-3.3}
Parameter-efficient fine-tuning}

The model can better adapt to the target task or data distribution by fine-tuning with a small number of task-specific samples, thereby improving performance and generation quality \citep{peft2,dreambooth}. However, in the few-shot font generation, the available data for fine-tuning is limited to only a few reference style images. Direct fine-tuning under such constraints can lead to overfitting. While the model may better adapt to the target style patterns, it can simultaneously overfit to the content patterns of the reference characters, resulting in content distortions, especially when generating unseen characters.

\noindent \textbf{Theoretical Analysis.} ALD-Font employs architecture-level disentanglement by injecting style information into the U-Net through cross-attention, as follows:
\begin{align}
&Q = W_Q\,h(z_{\mathrm{y}}^t,z_{\mathrm{x}}),
K = W_K\,s,
V = W_V\,s,\nonumber\\
&T=\frac{QK^\top}{\sqrt d},
A=\mathrm{softmax}(T),
O=AV,
\mathcal{L}=\mathcal{L}(O).
\nonumber
\end{align}
where \(h\) denotes modules before the cross-attention layer, $s$ represents style information extracted by CLIP, \(W_Q\), \(W_K\), and \(W_V\) are the projection matrices for the query, key, and value, respectively, \(d\) is the dimensionality of the hidden states, and \(\mathcal{L}\) represents the loss function associated with this layer. Gradients of the content and style conditions are:
\begin{align}
&G=\frac{\partial \mathcal L}{\partial O},
J_{\mathrm{softmax}}(A)=\mathrm{diag}(A)-AA^\top\nonumber\\
&\frac{\partial \mathcal L}{\partial z_{\mathrm{x}}}
=\Big(\frac{\partial h}{\partial z_{\mathrm{x}}}\Big)^{\!\top}
W_Q^{\!\top}\,
\frac{1}{\sqrt d}\;
J_{\mathrm{softmax}}^{\!\top}\!\!\left(
G V^\top
\right)
K,\\
&\frac{\partial \mathcal L}{\partial s}
=
\frac{1}{\sqrt d}\,W_K^{\!\top}\Big(J_{\mathrm{softmax}}^{\!\top}\big(G V^\top\big)\Big)^{\!\top} Q
\;+\;
W_V^{\!\top}A^{\!\top}G,
\end{align}
The gradient propagation of the content pathway depends on $K$ and $V$, since the attention mechanism couples the content query $Q$ with the style key $K$ and value $V$ through the $\mathrm{softmax}$ operation.
However, $K$ and $V$ serve only as weighting terms during backpropagation, influencing the updates of parameters such as $W_Q$ and $h$, while the parameters associated with $K$ and $V$ themselves remain unchanged.
Similarly, the optimization of the style channel $s$ depends on $Q$ only through attention weights.
Therefore, gradient propagation maintains separate optimization pathways for content and style.
When we fine-tune only style-related parameters, the content information participates in optimization but does not affect content-related parameters.

\begin{figure}[!htbp]

  \centering
  \includegraphics[width=.65\linewidth,height=0.45\textheight,keepaspectratio]{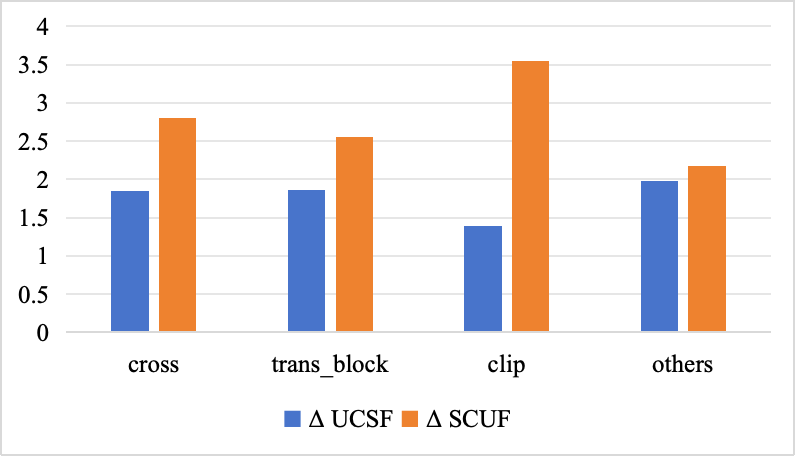}
  
\caption{Results of gradient analysis. Style-related parameters are more sensitive to unseen styles, while content-related parameters are more sensitive to unseen content.\label{fig:grad}}
\end{figure}

We further conduct a gradient analysis under three settings: Seen Chars Seen Font (SCSF), Unseen Chars Seen Font (UCSF), and Seen Chars Unseen Font (SCUF) based on the model trained in Section~{\protect\ref{Xsec7-3.2}}. We compute the gradients of all parameters within a batch of 128 samples under each of the three settings.
The gradient magnitude of a parameter group reflects its sensitivity to the current data distribution under gradient-based optimization. Specifically, the parameter update is determined by $\Delta \theta_i=-\eta\nabla_{\theta_i}\mathcal{L}$,
where a larger gradient norm indicates that the loss is more sensitive to this parameter group and that the model relies more on updating this group to adapt to the current distribution. Therefore, when a specific distribution shift is introduced, parameters with larger gradient responses are considered more relevant to adapting to that shift.
We categorize the parameters into four groups: (1) K and V projection matrices in Cross-Attention modules (KV), (2) Transformer Blocks (Trans\_block), (3) the last layer of CLIP (Clip), and (4) other parameters corresponding to content-related components (Others).
Although the KV are part of the Trans\_blocks, they directly receive style information and are therefore analyzed separately.
Taking SCSF as the baseline, we evaluate the sensitivity of each parameter to content and style variations.
For a given parameter \( x \), let the magnitude of the gradient under the SCSF setting be denoted as \( \text{G}(x) \), and the magnitude of the gradient under other settings as \( \text{G}_{*}(x) \).
We define their ratio \( \text{G}_{*}(x) / \text{G}(x) \) as the parameter sensitivity metric and report the mean sensitivity across each parameter group.
The results are shown in Fig.~{\ref{fig:grad}}.
Overall, the model exhibits higher sensitivity to unseen styles than to unseen characters.
KV, Trans\_block, and Clip exhibit higher sensitivity than Others to unseen styles in SCUF. Clip and KV show greater responses as they directly encode and propagate style information, whereas the Trans\_block plays a more auxiliary role in style adaptation.
In contrast, for unseen characters in UCSF, content-related components become more sensitive, while the CLIP module exhibits reduced gradient magnitude.
This gradient-based partition provides an empirical criterion for identifying functional roles of different parameter groups. The identified style-related parameters are consistent with the information pathway design of ALD-Font, where Transformer blocks and the CLIP projection layer directly receive or process style information.

Based on this observation, we treat all parameters that are more sensitive to style, including the Transformer blocks and the last layer of CLIP, as style-related parameters $\theta_{\mathrm{s}}$, and the remaining parameters as content-related parameters $\theta_{\mathrm{c}}$, corresponding to the red and yellow components in Fig.~{\ref{fig:overview}}, respectively. This partition is not strictly layer-wise: Transformer blocks occupy only part of the U-Net, while other parameters within the same layers, including up- and down-sampling modules, remain content-related.
These parameters mainly process the spatial feature stream and are therefore included in $\theta_{\mathrm{c}}$. We adopt a parameter-efficient fine-tuning strategy (PEFT) \citep{peft,mcc,dreambooth}, where only the style-related components $\theta_{\mathrm{s}}$ are updated. This enables the model to better adapt to the target style while preserving the content structure. Given the limited number of reference images, we aim to construct as many training pairs as possible. Given $N > 1$ reference images, we randomly select one as the target and choose subsets from the remaining $N - 1$ as reference inputs, resulting in $2^{N-1} - 1$ possible combinations. Thus, a total of $N(2^{N-1} - 1)$ unique training pairs can be constructed. During inference, we use all $N$ reference images to guide generation.
\begin{figure}[!htbp]

    \centering
    \includegraphics[width=.65\linewidth,height=0.45\textheight,keepaspectratio]{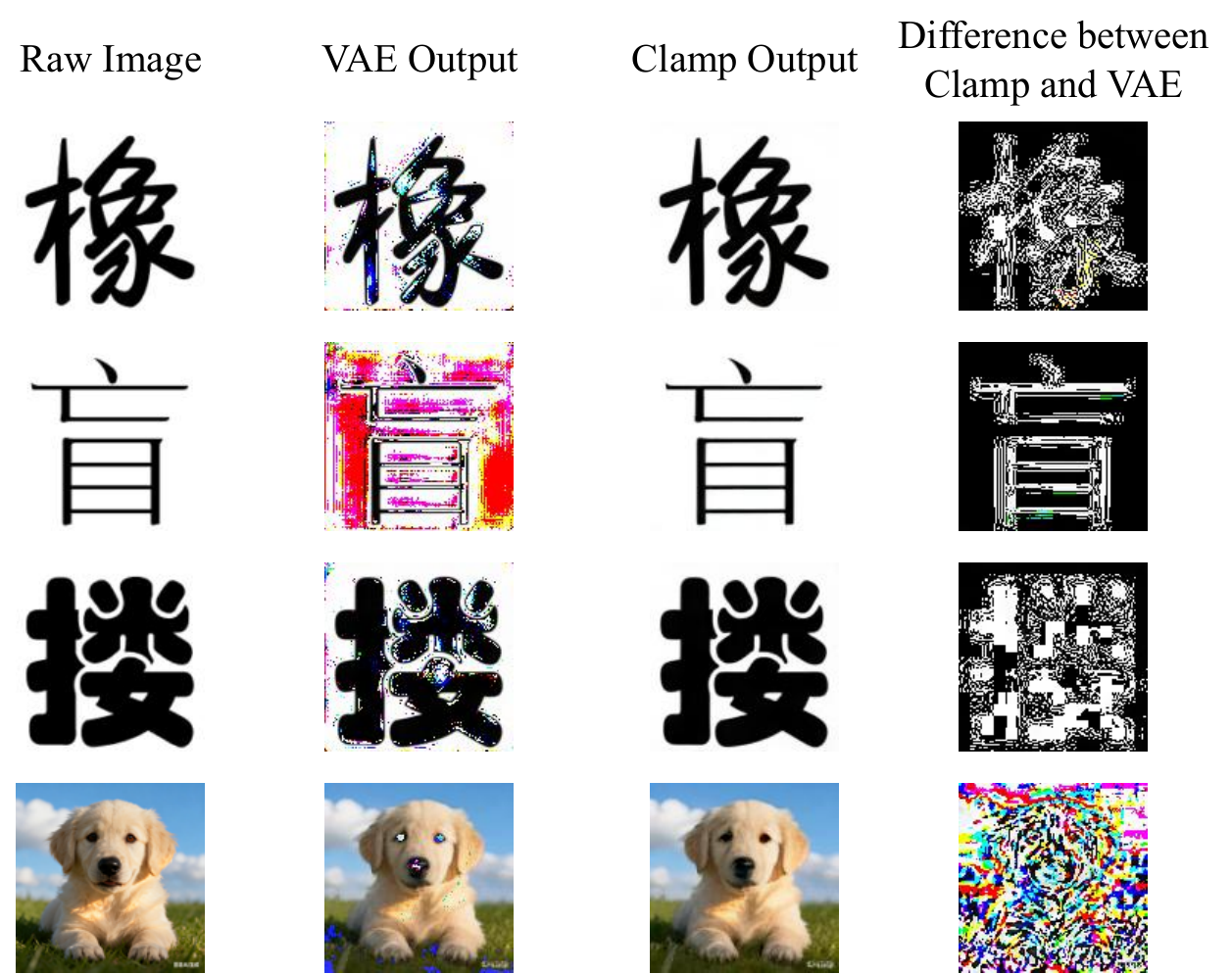}
    
\caption{Visualization of noise artifacts produced by the VAE reconstruction process. VAE output denotes the direct reconstruction produced by VAE encoding and decoding. Clamp output denotes the reconstruction after clipping pixel values to the range [0, 255]. The last column shows the difference between the clamped output and the original image. \label{fig:vae_test}}
\end{figure}
\begin{figure}[!htbp]

    \centering
    \includegraphics[width=.60\linewidth,height=0.45\textheight,keepaspectratio]{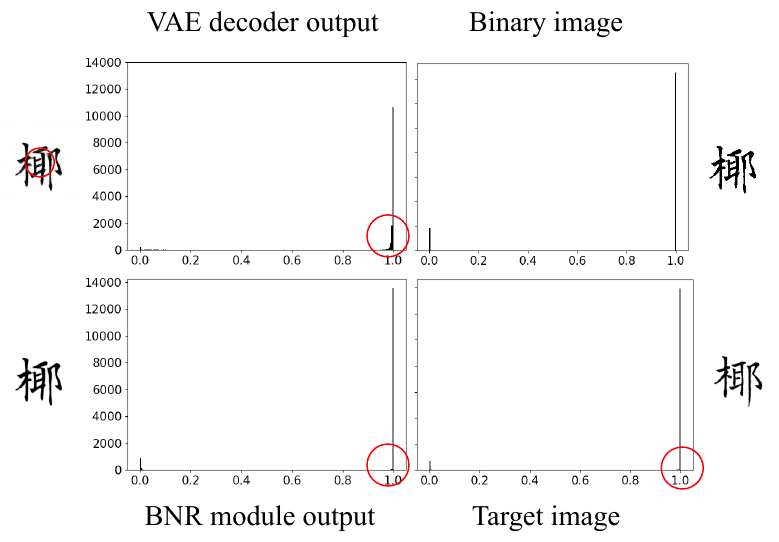}
    
\caption{The VAE output shows background noise in dense stroke areas, which is visible in the grayscale histogram. The BNR module effectively removes the noise.\label{fig:bnr}}
\end{figure}
\subsection{\label{Xsec9-3.4}
Background noise removal module}

Since Latent Diffusion Models rely on a VAE to transform between latent and pixel spaces, they inevitably introduce some noise during decoding \citep{vae,vae2013,vae2014}, as shown in Fig.~{\ref{fig:vae_test}}. Noise is often masked by rich color in regular text-to-image tasks. But it becomes visible in font generation, especially around dense strokes. As illustrated in Fig.~{\ref{fig:bnr}}, noise artifacts can be clearly observed in the densely stroked right part of the character when directly decoded from the VAE. This noise is reflected in the grayscale distribution: in the target image, pixel intensities are concentrated near 0 and 1, with only a small proportion of intermediate gray values.
In contrast, the VAE output shows a significant accumulation of pixels near 1, corresponding to the undesired noise regions.

We propose the Background Noise Removal (BNR) module to address the background noise present in generated Chinese characters. As shown in Fig.~{\ref{fig:bnr}}, although some noise exists, the overall grayscale tendency of the VAE output remains largely correct. Based on this observation, we convert the decoded image into a binary image and concatenate it with the SimSun source image $x$ along the channel dimension. This combined input is then fed into the BNR module.
The BNR module is implemented as a U-Net and outputs a noise-removed image:
\begin{align}
\tilde{y} = \text{BNR}(\text{Binarize}(\text{concat}(\mathcal{D}(\hat{z}_{\mathrm{y}}^0)), x)),
\end{align}
Since the BNR module is trained in pixel space, it allows us to incorporate edge and perceptual losses that are not feasible in latent space. Specifically, we apply a Sobel-based edge loss \citep{sobel} and a VGG perceptual loss \citep{vgg} for better stroke sharpness and visual quality. The overall loss function is defined as:
\begin{align}
\mathcal{L}_{\text{BNR}} = &\,\ell_1(\tilde{y}, y) + \lambda_1  \ell_1(\text{Sobel}(\tilde{y}),\text{Sobel}(y)) +  \lambda_2  \text{VGG}(\tilde{y}, y), \label{loss-bnr}
\end{align}
This setup enables the BNR module to effectively reduce background artifacts while preserving the content integrity of the generated characters.
We train the BNR module based on the trained ALD-Font.
It is worth noting that our ALD-Font adopts a training strategy that predicts $z_{\mathrm{y}}^0$, which means that at each diffusion time step, we can obtain an estimate of $\hat{z}_{\mathrm{y}}^0$ using Eq.~{\eqref{predict_x0}}. For clarity, we refer to ALD-Font as the default version with the BNR module.

\begin{table}[!htbp]%t1
\centering
\caption{Quantitative comparison under SCUF and UCUF settings across diverse evaluation metrics. Since Font-diff only supports SCUF, we do not report the results under UCUF. \textbf{SCSF}: Seen Chars Seen Font. \textbf{UCUF}: Unseen Chars Seen Font.}
\fitTable{%
\begin{tabular}{lllllllllllll}
\toprule
Model & \multicolumn{6}{l}{SCUF} & \multicolumn{6}{l}{UCUF} \\
& SSIM $\uparrow$  & LPIPS$\downarrow$ & FID$\downarrow$ & L1$\downarrow$ & Gray$\uparrow$ & OCR$\uparrow$
& SSIM$\uparrow$ & LPIPS$\downarrow$ & FID$\downarrow$ & L1 $\downarrow$ & Gray$\uparrow$ & OCR$\uparrow$ \\
\midrule
LF-Font                & 0.307 & 0.275 & 1.275  & 0.255 & 0.381 & 0.986 & 0.304 & 0.283 & 1.298  & 0.257 & 0.380 & 0.978 \\
MX-Font                & 0.319 & 0.256 & 0.671  & 0.249 & 0.285 & \textbf{1.000} & 0.321 & 0.259 & 0.689  & 0.246 & 0.283 & \textbf{0.999} \\
NTF               & 0.359 & 0.250 & 0.623  & 0.229 & 0.376 & 0.997 & 0.352 & 0.257 & 0.631  & 0.232 & 0.373 & 0.995 \\
DA-Font & 0.330 & 0.305 & 2.975 & 0.239 & 0.994 & 0.929 & 0.340 & 0.291 & 1.767 & 0.235 & 0.996 & 0.930 \\
Fontify (1shot) & 0.362 & 0.294 & 5.502 & 0.218 & 0.392 & 0.796 & 0.356 & 0.398 & 10.75 & 0.225 & 0.360 & 0.796 \\
Font-diff (1shot)  & 0.298 & 0.270 & 4.087  & 0.254 & 0.518 & 0.949 & -- & -- & -- & -- & -- & -- \\
FontDiffuser (1shot) & 0.347 & 0.265 & 2.770 & 0.237 & 0.984 & \textbf{1.000} & 0.341 & 0.270 & 2.621 & 0.240 & 0.983 & 0.999 \\
MSDFont           & 0.407 & 0.225 & 3.797 & 0.212 & 0.307 & 0.965 & 0.337 & 0.259 & 3.915 & 0.237 & 0.263 & 0.915 \\
MSDFont+FT        & 0.433 & 0.208 & 3.438 & 0.197 & 0.447 & 0.940 & 0.337 & 0.259 & 4.210 & 0.233 & 0.366 & 0.857 \\
\midrule
ALD-Font (1shot)  & 0.381 & 0.231 & 1.718 & 0.227 & 0.990 & 0.999 & 0.352 & 0.245 & 1.734 & 0.239 & 0.990 & 0.998 \\
ALD-Font          & 0.452 & 0.202 & 0.997 & 0.199 & 0.995 & 0.996 & 0.394 & 0.224 & 1.010 & 0.219 & 0.995 & 0.994 \\
ALD-Font+PEFT     & \textbf{0.505} & \textbf{0.175} & \textbf{0.521} & \textbf{0.171} & \textbf{0.998} & 0.991
& \textbf{0.436} & \textbf{0.202} & \textbf{0.558} & \textbf{0.196} & \textbf{0.997} & 0.989 \\
\bottomrule
\end{tabular}}
\label{tab:main-result}
\end{table}

\section{\label{Xsec10-4}
Experiment}

\subsection{\label{Xsec11-4.1}
Experimental settings}
\noindent \textbf{Implementation Details.} Following the implementation settings of \citep{msdfont}, all font images are resized to $128 \times 128 \times 3$, and the default number of reference images for few-shot generation is set to 8. The total number of diffusion steps is $T = 1000$. During training, each data sample consists of one SimSun source character image, one corresponding target-style character image, and 8 randomly selected non-target character images in the same target style as reference.
The weights in Eq.~{\eqref{loss-bnr}} are set as $\lambda_1 = 1$ and $\lambda_2 = 0.05$. During fine-tuning, we train the model separately for each font style for 80 epochs and report the results by averaging the evaluation metrics across all fonts.

\noindent \textbf{Datasets.} Following prior works \citep{msdfont, diff-font}, we collect 900 Chinese font styles from the Founder Type library, 840 for training (Seen Fonts) and the other 60 for evaluation (Unseen Fonts). We randomly sample 800 Chinese characters from each seen font as the training set. For testing, 200 characters are randomly selected from the training set as Seen Chars, and another 200 unseen characters as Unseen Chars. We evaluate in two scenarios: Seen Chars with Unseen Fonts (SCUF) and Unseen Chars with Unseen Fonts (UCUF).

\noindent \textbf{Evaluation Metrics.} We evaluate our method from both style and content perspectives. For style evaluation, we use several metrics: pixel-wise L1 loss, SSIM \citep{ssim} for structural similarity, LPIPS \citep{lpips} for perceptual quality, and FID \citep{fid} to assess distribution-level similarity. For content, inspired by Fig.~{\ref{fig:bnr}}, we introduce a metric, Gray, which computes the cosine similarity between the grayscale histograms of the generated and target images to reflect background noise and structural consistency. We evaluate character correctness using pretrained PaddleOCR~\citep{paddleocr} and a ResNet-based OCR model~\citep{resnet} trained on our dataset. A character is considered correct if recognized by either model.

\subsection{\label{Xsec12-4.2}
Comparison with state-of-the-art methods}

We compare ALD-Font with various methods, including LF-Font \citep{lf}, MX-Font \citep{mx}, NTF \citep{ntf}, Font-diff \citep{diff-font}, DA-Font \citep{dafont}, Fontify \citep{fontify}, FontDiffuser \citep{fontdiffuser}, and MSDFont \citep{msdfont}.  Fontify, FontDiffuser, and Font-diff are one-shot methods. We also evaluate ALD-Font in the one-shot setting without retraining by randomly selecting a reference image as input.
For fairness, all methods are trained on our collected dataset following their official implementations.

As shown in Table~{\ref{tab:main-result}}, under both settings, ALD-Font consistently achieves the best performance compared to other methods.
This demonstrates that the asymmetric architecture enables better structure preservation and perceptual quality.
Building on architecture-level disentanglement, ALD-Font+PEFT greatly improves style consistency, achieving the best performance across all style metrics. For content-related metrics, Gray reaches 0.998, and OCR remains high at 0.991. Although content performance slightly declines, style performance improves significantly, consistent with our expected behavior under disentanglement.
MX-Font obtains a high OCR score mainly because its content pathway is strong. It extracts local content features from the target character image, which explicitly provides stroke topology, component layout, and global character structure. However, its style features are averaged over multiple references before being fused with the content features, which may blur fine-grained style details.
The OCR score of ALD-Font+PEFT is slightly lower than that of ALD-Font (1-shot) because PEFT encourages the model to capture more target-style characteristics and stroke details. These style-specific deformations may slightly reduce OCR confidence. A similar trend is observed when increasing the number of reference images from one to multiple references. Nevertheless, the OCR scores of ALD-Font and ALD-Font+PEFT remain high and within an acceptable range, while their style-related metrics are significantly improved. This trade-off is expected because PEFT prioritizes style adaptation over preserving the canonical glyph shape.
Under the UCUF setting, results are similar to those observed in SCUF, with ALD-Font+PEFT achieving superior overall performance compared to existing approaches. Since PEFT only optimizes style information and lacks adaptation to content, the improvement is less significant than in the SCUF setting.
For comparison, we also fine-tune MSDFont (MSDFont+FT). In the SCUF setting, although MSDFont+FT shows improved style metrics, its content performance drops. In the UCUF setting, the overall performance even degrades, highlighting the importance of architecture-level content--style disentanglement, especially for OCR, indicating that the model is overfitted to the content.

\subsection{\label{Xsec13-4.3}
Visualization analysis}

\begin{figure}[!htbp]

  \centering
  \includegraphics[width=1.0\linewidth,height=0.68\textheight,keepaspectratio]{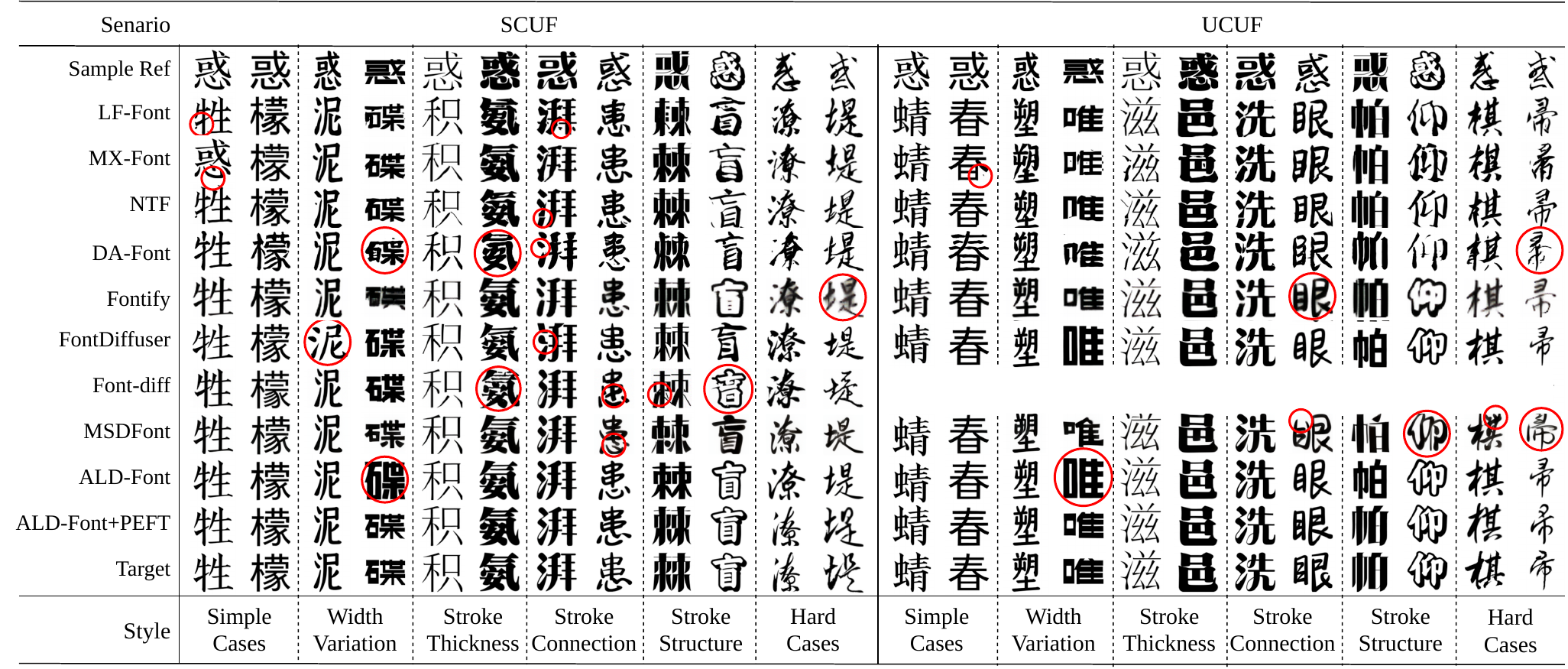}
  
\caption{Visualization of Chinese character generation results under SCUF and UCUF settings, illustrating two fonts for each of six representative style characteristics. Problematic generations are highlighted with red circles.\label{fig:visualize}}
\end{figure}

To better understand the effectiveness of our method, we visualize the generation results in Fig.~{\ref{fig:visualize}}.
It can be seen that all models accurately capture both the content and style of the target font in simple cases, while LF-Font and MX-Font show background noise artifacts.
FontDiffuser fails to handle narrow-width fonts, while ALD-Font struggles with flat-shaped fonts.
However, after applying PEFT, the generated fonts closely match the target style. Although all methods perform well on thin fonts, LF-Font, NTF, DA-Font, FontDiffuser, and Font-diff exhibit noise in complex stroke regions when generating thicker fonts.
In terms of stroke connection, only ALD-Font+PEFT successfully reproduces the correct connectivity consistent with the target font. Regarding stroke structure, the rightmost example shows a font with shadowed strokes. Fontify and Font-diff capture the stylistic shadowing but fail to produce the correct image. ALD-Font generates the correct character structure, though the shadow effect is less pronounced. ALD-Font+PEFT achieves the best balance, accurately preserving both structure and stylistic detail.
The most challenging cases involve semi-cursive and cursive scripts, which inherently deviate from standard stroke structures. Most baseline methods prioritize structural fidelity, thereby failing to imitate the target style closely. ALD-Font+PEFT generates outputs that best resemble the style of the reference. Overall, ALD-Font+PEFT achieves the best generation results.

\begin{figure}[!htbp]

\centering
  \includegraphics[width=.60\linewidth,height=0.45\textheight,keepaspectratio]{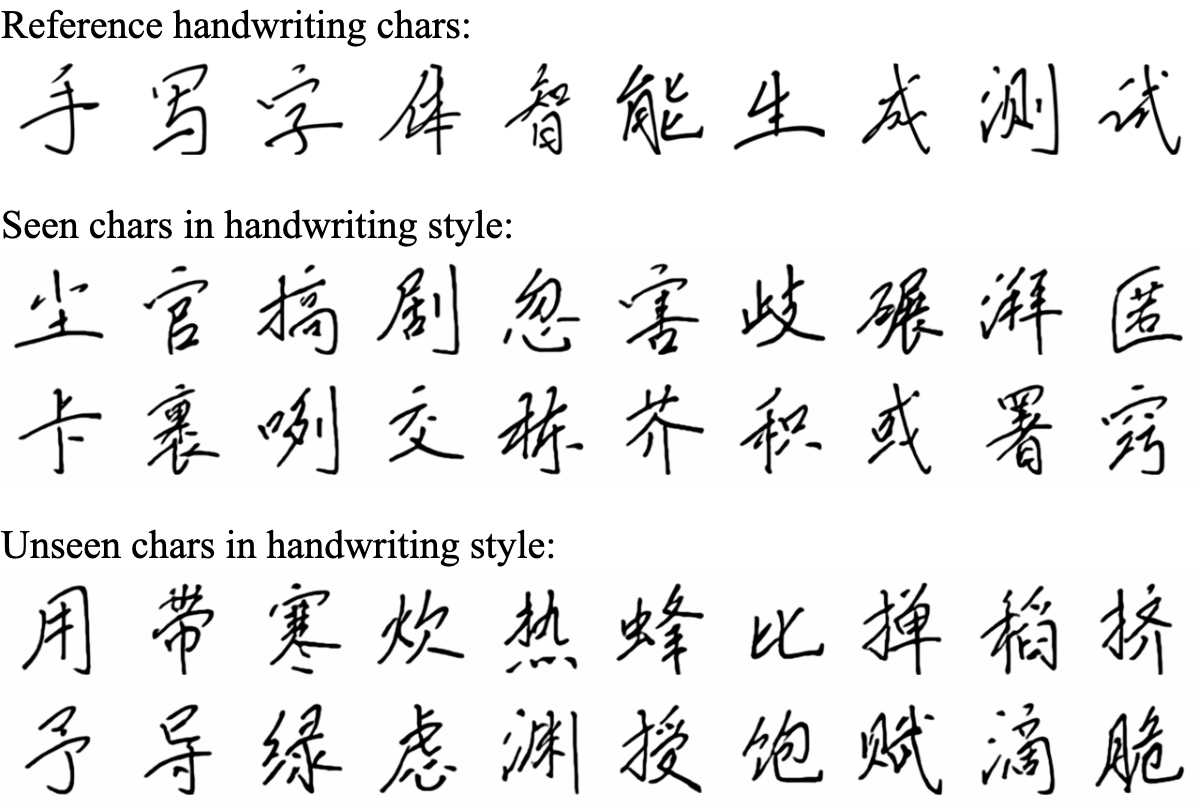}
  
\caption{Visualization of handwritten style test.\label{fig:handwritten}}
\end{figure}

We further evaluate ALD-Font+PEFT using handwritten characters as references, which serve as Out-Of-Distribution (OOD) style data due to their higher variability compared to printed fonts. Results are reported on both seen and unseen characters. As shown in Fig.~{\ref{fig:handwritten}}, the results demonstrate that ALD-Font+PEFT effectively captures the reference style, producing characters visually consistent with the handwritten examples, which indicates strong style fitting capability.

\subsection{\label{Xsec14-4.4}
Ablation study}

In this section, we conduct ablation studies on ALD-Font to investigate the effectiveness of the fine-tuning strategy, BNR module, and image encoder size. All experiments are conducted under the UCUF setting.

\subsubsection{\label{Xsec15-4.4.1}
Effect of fine-tuning strategy}

To validate the effectiveness of the proposed PEFT strategy, we compare different fine-tuning strategies, including updating the last CLIP layer (CLIP), adjusting the K/V projections in cross-attention (KV), tuning all parameters (ALL), applying LoRA \citep{lora} to linear layers with rank 8 (LoRA), and the proposed PEFT strategy. The results are shown in Table~{\ref{tab:ablation}}.

\begin{table}[!htbp]
\centering
\caption{Effect of different fine-tuning strategies in ALD-Font. ``CLIP'' tunes the last CLIP layer, ``KV'' tunes the K/V projections, ``ALL'' tunes all parameters, ``LoRA'' applies LoRA to linear layers, and ``PEFT'' applies parameter-efficient fine-tuning.\label{tab:ablation}}
\fitTable{%
\begin{tabular}{llllllll}
\toprule
Strategy & BNR & SSIM$\uparrow$ & LPIPS$\downarrow$ & FID$\downarrow$ & L1$\downarrow$ & Gray$\uparrow$ & OCR$\uparrow$\\
\midrule
No & $\times$ & 0.377 & 0.222 & 2.189 & 0.220 & 0.896 & 0.993\\
CLIP & $\times$ & 0.376 & 0.222 & 2.187 & 0.220 & 0.896 & 0.992\\
KV & $\times$ & 0.390 & 0.214 & 1.781 & 0.212 & 0.898 & 0.994\\
ALL & $\times$ & 0.388 & 0.210 & 2.164 & 0.210 & 0.510 & 0.973\\
LoRA & $\times$ & 0.410 & 0.206 & 1.711 & 0.202 & 0.893 & 0.976\\
PEFT & $\times$ & 0.423 & \textbf{0.202} & 1.719 & \textbf{0.196} & 0.779 & 0.986\\
\midrule
No & $\checkmark$ & 0.394 & 0.224 & 1.010 & 0.219 & 0.995 & 0.994\\
CLIP & $\checkmark$ & 0.395 & 0.224 & 1.035 & 0.219 & 0.995 & 0.994\\
KV & $\checkmark$ & 0.410 & 0.217 & 0.735 & 0.211 & \textbf{0.997} & \textbf{0.996}\\
ALL & $\checkmark$ & 0.405 & 0.215 & 0.642 & 0.208 & \textbf{0.997} & 0.984\\
LoRA & $\checkmark$ & 0.429 & 0.209 & 0.865 & 0.200 & 0.996 & 0.981\\
PEFT & $\checkmark$ & \textbf{0.436} & \textbf{0.202} & \textbf{0.558} & \textbf{0.196} & \textbf{0.997} & 0.989\\
\bottomrule
\end{tabular}}
\end{table}

The CLIP strategy provides only limited improvement over the baseline, indicating that adapting the high-level style representation alone is insufficient. In contrast, KV, LoRA, ALL, and PEFT improve style adaptation. KV achieves moderate style enhancement with limited content degradation, while ALL causes larger content degradation due to updating content-related parameters. Although LoRA achieves competitive performance, the architecture-level disentanglement in ALD-Font enables PEFT to achieve better style adaptation under few-shot settings. These results further validate the gradient-based parameter grouping, where updating style-sensitive parameters improves style fidelity while preserving content-related performance.

\subsubsection{\label{Xsec16-4.4.2}
Effect and sensitivity analysis of BNR module}

\begin{figure}[!htbp]

    \centering
    \includegraphics[width=.65\linewidth,height=0.45\textheight,keepaspectratio]{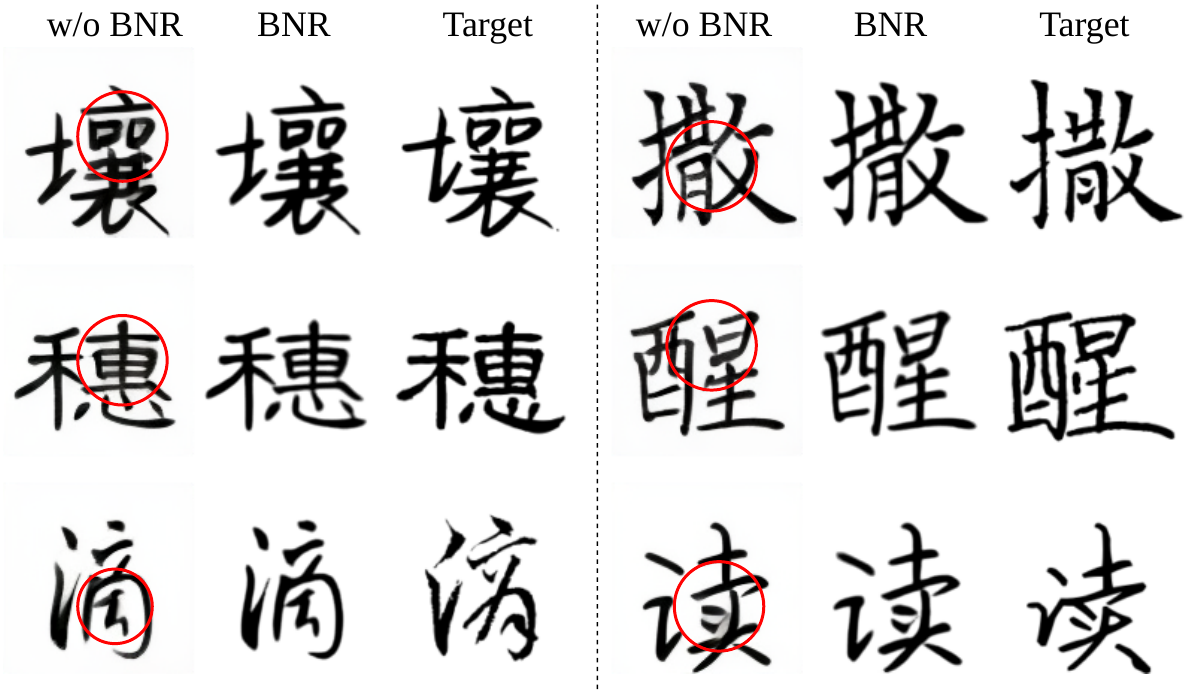}
    
\caption{Visual comparison of ALD-Font with and without BNR. BNR removes background artifacts and noise in complex stroke regions, but does not directly correct style-level discrepancies in the generated glyphs.\label{fig:bnr-sample}}
\end{figure}

To evaluate the effectiveness of the proposed BNR module, we compare ALD-Font with and without BNR. As shown in Table~{\ref{tab:ablation}}, BNR consistently improves content-related metrics by removing VAE-induced background artifacts, while the resulting improvement in content consistency also indirectly benefits style-related metrics. Fig.~{\ref{fig:bnr-sample}} further visualizes this effect: BNR effectively suppresses noise in complex stroke regions and across the overall background. However, the last row shows that BNR only refines the output produced by ALD-Font and cannot directly correct style-level discrepancies.

To further analyze the effect of BNR at the stroke level, we conduct a connected-component analysis on the overall UCUF test set and additionally examine a complex cursive font (DHTCSJW) and  handwritten fonts. Specifically, we first binarize the generated glyphs and then extract their connected components. We report the median and interquartile range (Q1, Q3) of the total number of connected components (\textit{Comp.}) and the number of small components with an area no larger than 8~pixels (\textit{Small Comp.}); the latter mainly reflects isolated noise points and local artifacts. As shown in Table~{\ref{tab:bnr-component}}, BNR substantially reduces Small Comp. across all settings. For cursive and the overall UCUF set, the median Small Comp. decreases to 0, while the corresponding Comp. values become close to those of the target glyphs. For handwritten fonts, although Comp. slightly increases because of their highly flexible and irregular stroke structures, Small Comp. decreases markedly from 6.0 to 1.0. These results indicate that BNR mainly removes isolated noise and artifacts in dense or complex stroke regions rather than modifying the main stroke connections. The remaining structural deviations in cursive and handwritten fonts are primarily caused by the difficulty of generating their highly variable glyph structures, while BNR only performs pixel-level refinement and does not alter the overall stroke layout.

\begin{table}[!htbp]
\centering
\caption{Connected-component analysis before and after applying BNR. Values are reported as median [Q1, Q3]. Comp. denotes the number of connected components, while Small Comp. denotes components with an area no larger than 8~pixels.\label{tab:bnr-component}}
\fitTable{%
\begin{tabular}{llll}
\toprule
Scope & Group & Comp. & Small Comp. \\
\midrule
cursive
& w/o BNR & 15.0 [11.0, 19.0] & 7.0 [5.0, 11.0] \\
& BNR & 3.0 [2.0, 4.0] & 0.0 [0.0, 0.0] \\
& Target & 4.0 [3.0, 7.0] & 1.0 [0.0, 2.0] \\
\midrule
Overall UCUF
& w/o BNR & 11.0 [6.0, 17.0] & 5.0 [2.0, 9.0] \\
& BNR & 3.0 [2.0, 5.0] & 0.0 [0.0, 0.0] \\
& Target & 4.0 [2.0, 5.0] & 0.0 [0.0, 0.0] \\
\midrule
Handwritten
& w/o BNR & 8.0 [7.0, 11.0] & 6.0 [4.0, 8.0]\\
& BNR & 9.0 [5.0, 12.0] & 1.0 [0.0, 4.0]\\
\bottomrule
\end{tabular}}
\end{table}

\begin{figure}[!htbp]

    \centering
    \includegraphics[width=.90\linewidth,height=0.68\textheight,keepaspectratio]{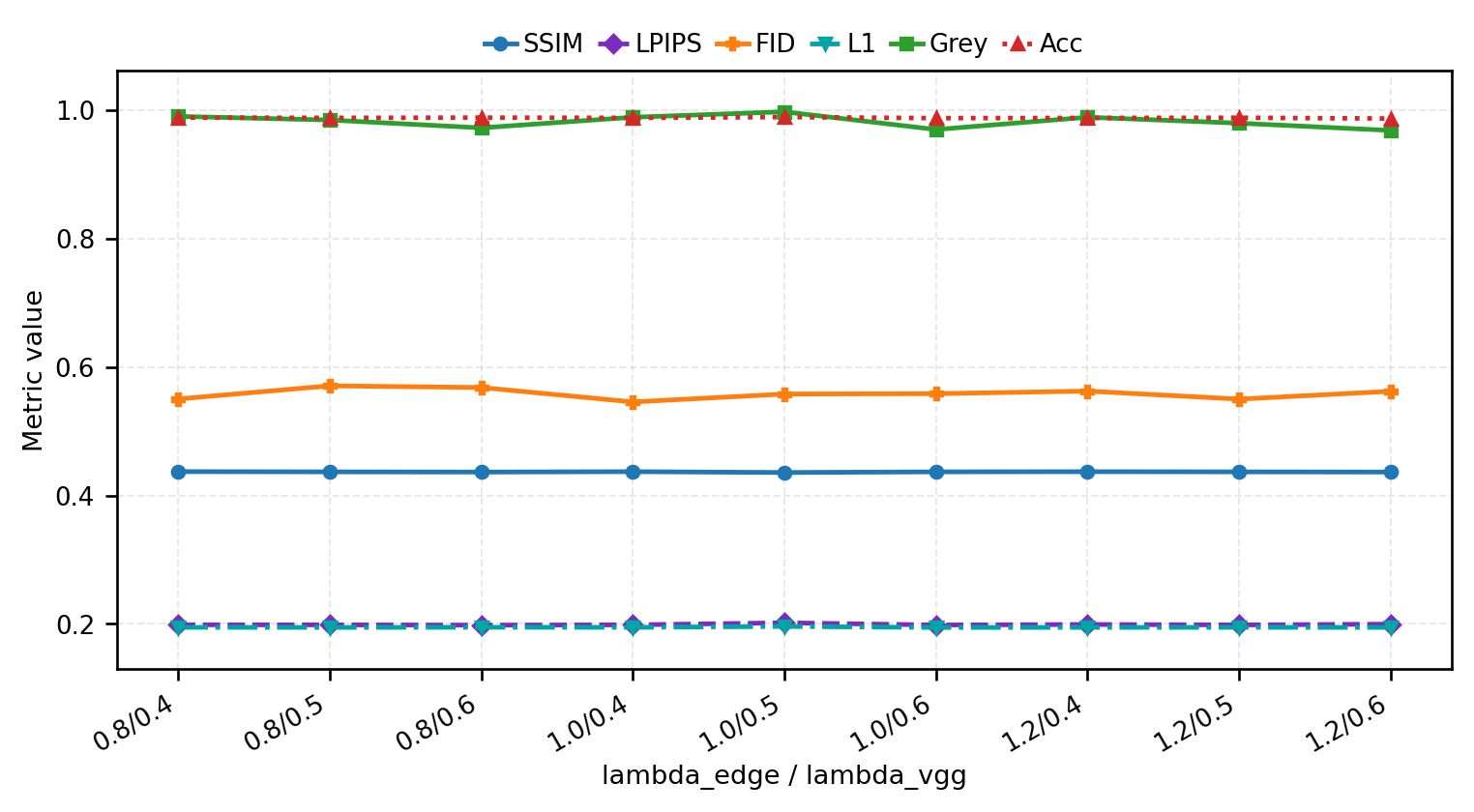}
    
\caption{Hyperparameter sensitivity analysis of the BNR loss weights. The edge-loss weight $\lambda_1$ and VGG-loss weight $\lambda_2$ are varied over nine combinations.\label{fig:bnr-sensitivity}}
\end{figure}

Furthermore, we conduct a hyperparameter sensitivity analysis on the BNR loss weights in Eq.~{\eqref{loss-bnr}}. Specifically, we evaluate nine combinations of the edge-loss weight $\lambda_1$ and the VGG-loss weight $\lambda_2$ around the default setting. As shown in Fig.~{\ref{fig:bnr-sensitivity}}, all evaluation metrics, including SSIM, LPIPS, FID, $\ell_1$, Gray, and OCR, remain stable across different weight combinations, with only minor fluctuations. These results indicate that BNR is insensitive to moderate variations in the loss weights.

\subsubsection{\label{Xsec17-4.4.3}
Effect of image encoder size}

The original experiments adopt ChineseCLIP-L14 due to its stronger feature representation capability and better generation quality. To investigate whether ALD-Font relies on a large image encoder, we replace ChineseCLIP-L14 with a smaller ChineseCLIP-B16 while keeping other components unchanged. The results are shown in Table~{\ref{tab:b16-result}}.

\begin{table}[!htbp]
\centering
\caption{Ablation study on the image encoder size.\label{tab:b16-result}}
\fitTable{%
\begin{tabular}{lllllll}
\toprule
Setting & SSIM$\uparrow$ & LPIPS$\downarrow$ & FID$\downarrow$ & L1$\downarrow$ & Gray$\uparrow$ & OCR$\uparrow$\\
\midrule
SCUF (1-shot) & 0.393 & 0.232 & 1.879 & 0.218 & 0.992 & 0.995\\
SCUF & 0.431 & 0.207 & 1.192 & 0.203 & 0.995 & 0.994\\
SCUF+PEFT & 0.473 & 0.185 & 0.526 & 0.182 & 0.998 & 0.986\\
UCUF(1-shot) & 0.364 & 0.248 & 1.905 & 0.230 & 0.992 & 0.993\\
UCUF & 0.387 & 0.228 & 1.226 & 0.220 & 0.995 & 0.992\\
UCUF+PEFT & 0.418 & 0.209 & 0.571 & 0.203 & 0.997 & 0.981\\
\bottomrule
\end{tabular}}
\end{table}

As shown in Table~{\ref{tab:b16-result}}, ALD-Font-B16 achieves competitive performance with a smaller backbone under both SCUF and UCUF settings. Although the B16 version slightly underperforms the L14 version due to its reduced representation capability, it still achieves strong performance, demonstrating that the effectiveness of ALD-Font does not solely rely on a larger CLIP backbone.

\subsection{\label{Xsec18-4.5}
Efficiency analysis}

We evaluate the computational cost of ALD-Font from two perspectives: the parameter-efficient adaptation process and the overall model complexity.

\subsubsection{\label{Xsec19-4.5.1}
Efficiency of parameter-efficient fine-tuning}

We first evaluate the efficiency of different fine-tuning strategies for adapting a single font style. As shown in Table~{\ref{tab:finetune_time}}, although different strategies vary substantially in the number of trainable parameters, their actual fine-tuning times remain similar. This is mainly because, under the few-shot setting, only a limited number of reference images are available, and the major computational cost comes from data processing and latent encoding rather than parameter updates. Consequently, PEFT achieves a favorable trade-off between efficiency and adaptation capability. Additionally, the BNR module is lightweight and introduces only 0.0033~s of additional inference time per image, resulting in negligible overhead during deployment.

\begin{table}[!htbp]
\centering
\caption{Comparison of different fine-tuning strategies in terms of trainable parameters and fine-tuning time on a single NVIDIA 4090 GPU.\label{tab:finetune_time}}
\fitTable{%
\begin{tabular}{lll}
\toprule
Strategy & Params (M) & Time (s)\\
\midrule
CLIP & 0.79 & 69.11\\
KV & 2.36 & 68.09\\
ALL & 34.67 & 74.48\\
LoRA & 0.21 & 66.61\\
PEFT & 7.30 & 67.57\\
\bottomrule
\end{tabular}}
\end{table}

\subsubsection{\label{Xsec20-4.5.2}
Overall model complexity}

To provide a comprehensive comparison with existing font generation methods, we further report the inference parameters, trainable parameters, GPU training time, and prior information used by different methods in Table~{\ref{tab:efficiency}}. Although ALD-Font introduces an additional BNR module, the overall overhead remains limited. The BNR module contains only 7.71M parameters and requires 32 GPU-hours for training. The relatively large inference parameter size of ALD-Font mainly comes from the pretrained latent diffusion backbone and CLIP encoder.
Some GAN-based methods have more trainable parameters than inference parameters because the discriminator is discarded after training. Moreover, although ALD-Font utilizes pretrained encoders, existing methods also rely on different forms of prior information, such as manually annotated component priors, stroke priors, pretrained visual representations, or latent diffusion priors, as illustrated in Section~{\protect\ref{Xsec3-2.1}}. Font generation places substantially greater emphasis on glyph quality than on processing speed. Considering the substantial performance gains achieved by ALD-Font, the additional computational cost remains acceptable.

\begin{table}[!htbp]
\centering
\caption{Comparison of model parameters and training costs. Infer. Params denotes the parameters used during inference, while Train. Params denotes the parameters updated during training. For GAN-based methods, trainable parameters may be larger because the discriminator is trained but discarded during inference.\label{tab:efficiency}}
\fitTable{%
\begin{tabular}{llll}
\toprule
Method & Infer. Params & Train. Params & GPU-h\\
\midrule
LF-Font & 7.98M & 27.05M & 92\\
MX-Font & 22.76M & 42.27M & 158\\
NTF & 9.08M & 19.55M & 50\\
DA-Font & 9.29M & 18.78M & 173\\
Fontify & 154.38M & 139.08M & 332\\
Font-diff & 132.50M & 123.21M & 171\\
FontDiffuser & 100.49M & 100.49M & 50\\
MSDFont & 235.30M & 101.94M & 446\\
\midrule
ALD-Font (L14) & 421.52M & 33.89M & 248\\
ALD-Font (B16) & 203.93M & 33.89M & 215\\
BNR Module & 7.71M & 7.71M & 32\\
\bottomrule
\end{tabular}}
\end{table}

\subsection{\label{Xsec21-4.6}
English letters generation evaluation}

\begin{table}[!htbp]
\centering
\caption{Quantitative comparison of English letters. ALD-Font achieves better scores in all evaluation metrics.\label{tab:res-eng}}
\fitTable{%
\begin{tabular}{lllllll}
\toprule
Method & SSIM$\uparrow$  & LPIPS$\downarrow$ & Fid$\downarrow$ & L1$\downarrow$ & Gray$\uparrow$ & OCR$\uparrow$ \\
\midrule
FontDiffuser & 0.368 & 0.456 & 3.920 & 0.278 & 0.976 & 0.288 \\
ALD-Font & \textbf{0.404} & \textbf{0.385} & \textbf{2.434} & \textbf{0.256} & \textbf{0.982} & \textbf{0.461} \\
\bottomrule
\end{tabular}}
\end{table}
\begin{figure}[!htbp]

  \centering
  \includegraphics[width=.65\linewidth,height=0.45\textheight,keepaspectratio]{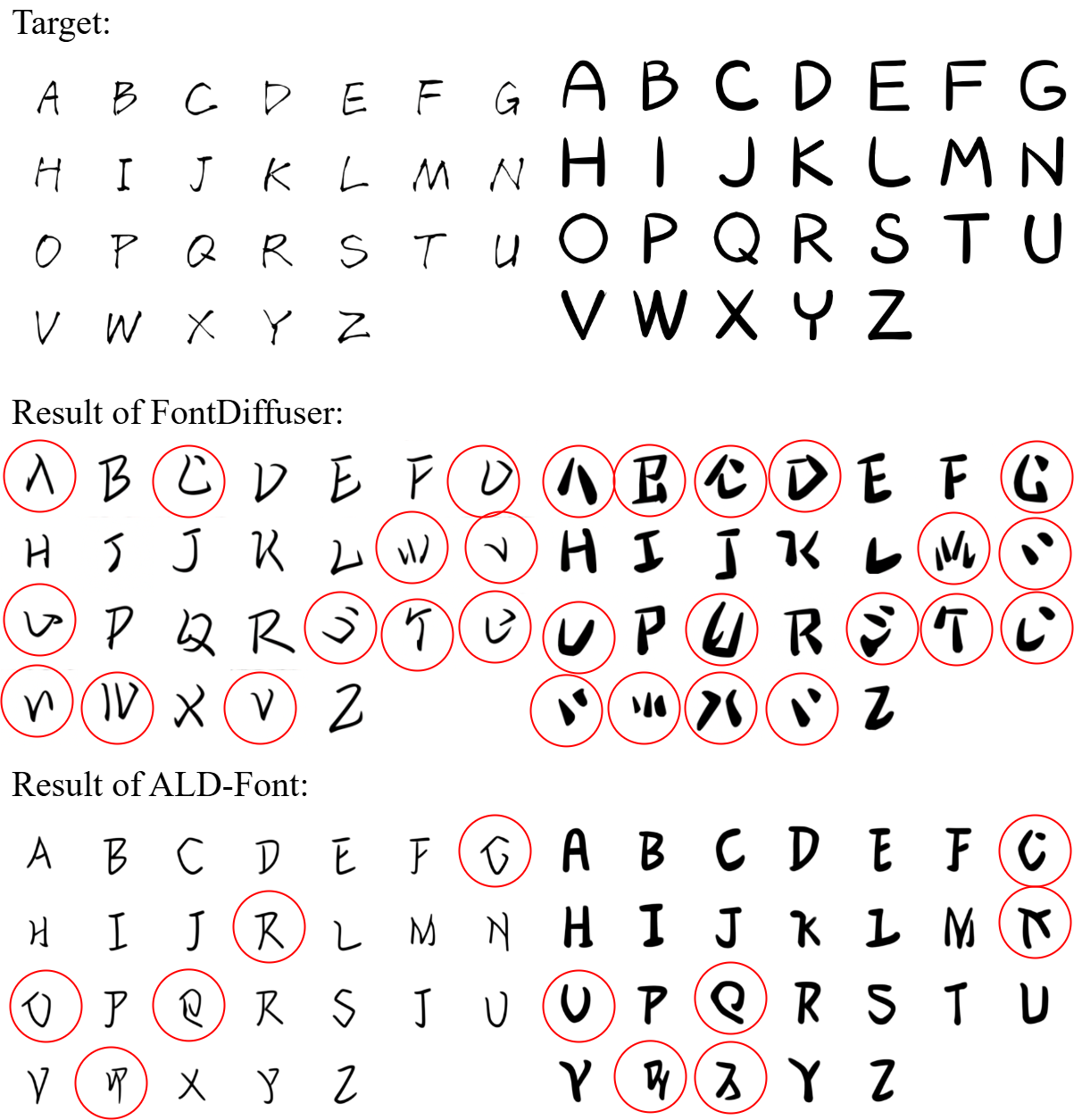}
  
\caption{Visualization of the English letters test. Both methods struggle to generate accurate letters due to the lack of alphabet training data, but ALD-Font produces better results than FontDiffuser.\label{fig:english-char}}
\end{figure}

To evaluate the content robustness of ALD-Font, we compare it with FontDiffuser \citep{fontdiffuser}, which demonstrates the best overall structural performance among prior methods, on the task of generating English letters. Both models are trained only on Chinese data and tasked with generating structurally different English letters, making it a strong Out-Of-Distribution test of content generalization. The high-scoring method MX \citep{mx}, which relies on stroke decomposition of Chinese characters, does not support English letters and is therefore excluded from the comparison. Since the references are Chinese characters, this scenario presents an additional challenge for maintaining structural consistency.

We present a quantitative comparison between ALD-Font and FontDiffuser \citep{fontdiffuser} on English letter generation in Table~{\ref{tab:res-eng}}. Results show that although neither method is specifically designed for English generation, ALD-Font achieves better performance than FontDiffuser in both content accuracy and style consistency. Visualization results are shown in Fig.~{\ref{fig:english-char}}. Although both models struggle with English letter generation, ALD-Font clearly produces more structurally accurate results, as evidenced by letters such as A, M, S, U, and Y. This supports our hypothesis that using the source content image as a content template provides stronger robustness in preserving character structure.

\section{\label{Xsec22-5}
Conclusion}

In this paper, we introduce ALD-Font, an architecture-level disentangled diffusion model for few-shot Chinese font generation. By using content as the backbone and style as guidance, our model achieves effective architecture-level disentanglement.
Meanwhile, a parameter-efficient fine-tuning strategy enables rapid adaptation to new styles without overfitting to content patterns of reference images in the few-shot setting, and the BNR module is introduced to further enhance visual clarity.
Extensive experiments demonstrate that ALD-Font preserves content integrity and significantly improves style fidelity over existing diffusion-based methods.
We hope this work inspires future research on architecture-level disentanglement for controllable font generation.

\section*{Acknowledgements}

This work was partially supported by the National Natural Science Foundation of China (Grant No. 62276127) and the Fundamental and Interdisciplinary Disciplines Breakthrough Plan of the Ministry of Education of China (No. JYB2025XDXM118), and the ``111 Center'' (No. B26023).

\end{document}